\UseRawInputEncoding
\documentclass[journal=jacsat,manuscript=article]{achemso}
\pdfoutput=1
\usepackage[version=4]{mhchem}
\usepackage{textcomp}

\usepackage{chemformula} 
\usepackage[T1]{fontenc} 
\usepackage{bm}
\usepackage{amsfonts}
\usepackage{amsmath}
\usepackage{natbib}
\usepackage[english]{babel}
\usepackage{graphicx}
\usepackage{url}
\usepackage{bm}
\usepackage{hyperref}
\usepackage{mhchem}
\usepackage[absolute,overlay]{textpos}
\usepackage{wrapfig}
\usepackage{multicol}
\usepackage{nicematrix}
\usepackage{pgfplots}
\usepackage{tikz}
\usepackage{booktabs}
\usepackage{listings}
\usepackage{xcolor}
\usepackage{epstopdf}
\usepackage{multirow}
\usepackage[normalem]{ulem}

\newcommand{\celsius}{$^{\circ}$C}
\newtheorem{problemx}{Problem (POD prediction under multi-modal physiological data)}

\newtheorem{definition}{Definition(TrendLoss)}

\author{ Bingxu Wang}

\author{Min Ge}

\author{ Kunzhi Cai}

\author{ Yuqi Zhang}

\author{Zeyi Zhou}
\affiliation[Nanjing Drum Tower Hospital]{Department of Thoracic and Cardiovascular Surgery, The Affiliated Drum Tower Hospital of Nanjing University Medical School, Kuang Yaming Honors School, Nanjing University, Nanjing 210023, China
}

\author{Wenjiao Li}
\affiliation[Nanjing Drum Tower Hospital]{Department of Thoracic and Cardiovascular Surgery, The Affiliated Drum Tower Hospital of Nanjing University Medical School, Kuang Yaming Honors School, Nanjing University, Nanjing 210023, China
}
\author{ Yachong Guo}
\email{yguo@nju.edu.cn}
\affiliation[Nanjing Drum Tower Hospital]{Department of Thoracic and Cardiovascular Surgery, The Affiliated Drum Tower Hospital of Nanjing University Medical School, Kuang Yaming Honors School, Nanjing University, Nanjing 210023, China
}

\author{Wei Wang}
\email{wangwei@nju.edu.cn}
\affiliation[Nanjing University]{National Laboratory of Solid State Microstructure, Department of Physics, Nanjing University, Nanjing 210093, China}

\author{Qing Zhou}
\affiliation[Nanjing Drum Tower Hospital]{Department of Thoracic and Cardiovascular Surgery, The Affiliated Drum Tower Hospital of Nanjing University Medical School, Kuang Yaming Honors School, Nanjing University, Nanjing 210023, China
}
\email{zhouqing@njglyy.com}

\title[Transformer representation learning is necessary for dynamic multi-modal physiological data on small-cohort patients]
  {Transformer representation learning is necessary for dynamic multi-modal physiological data on small-cohort patients}

\begin{document}







\begin{abstract}
  Postoperative delirium (POD), a severe neuropsychiatric complication affecting nearly 50\% of high-risk surgical patients, is defined as an acute disorder of attention and cognition, It remains significantly underdiagnosed in the intensive care units (ICUs) due to subjective monitoring methods. Early and accurate diagnosis of POD is critical and achievable. Here, we propose a POD prediction framework comprising a Transformer representation model followed by traditional machine learning algorithms. Our approaches utilizes multi-modal physiological data, including amplitude-integrated electroencephalography (aEEG), vital signs, electrocardiographic monitor data as well as  hemodynamic parameters. We curated the first multi-modal POD dataset encompassing two patient types and evaluated the various Transformer architectures for representation learning. Empirical results indicate a consistent improvements of sensitivity and Youden index in patient TYPE I using Transformer representations, particularly our fusion adaptation of Pathformer. By enabling effective delirium diagnosis from postoperative day 1 to 3, our extensive experimental findings emphasize the potential of multi-modal physiological data and highlight the necessity of representation learning via multi-modal Transformer architecture in clinical diagnosis.
\end{abstract}

\section{Introduction}
Postoperative delirium(POD), a prevalent acute neuropsychiatric syndrome\cite{RN6,10.1001/jama.2017.12067}, affects more than 50\% of surgical patients and significantly elevates morbidity and mortality risks\cite{10.1001/jamanetworkopen.2023.37239}. Early identification is crucial yet challenging\cite{RN11}, primarily due to subjective assessment criteria and incomplete understanding of underlying pathophysiological mechanisms\cite{Gamberale2021-jc}.  Risk factors previously identified include age, functional impairment, individual lifestyle habits, and surgical procedures\cite{JXYM4361, RN12, Bilotta2021, Dogan2023-kr, Kiani2022-nm}. While early predictive methods using machine learning models such as Logistic Regression, Random Forest, and Support Vector Machine(SVM) have demonstrated promising results\cite{Lee2023, Yoshimura2024-cg, Song2024}, these approaches primarily utilize static data collected before or after surgeries, neglecting dynamic  physiological fluctuations. Emerging evidence suggests the potential of sequential physiological signals\cite{Palanca2017-yn, Yang2016-jz, Fuest2022-ff}, including electroencephalography(EEG), for early POD detection, yet comprehensively representation learning remains underexplored\cite{Wong2018-bi}.

Advancements in ICU monitoring technology have enabled continuous, multi-modal data collection\cite{Boss2022-gf}, creating opportunities for improved early diagnosis and real-time clinical decision support\cite{Wang2023-wj, hayat2021dynamicmultimodalphenotypingusing}. However, the inherent heterogeneity and complexity of physiological data present analytical challenges\cite{Li2022, Tao2024, shao2024multimodalphysiologicalsignalsrepresentation,mordacq2024adaptmultimodallearningdetecting}. Here, we propose a novel Transformer representation learning framework that effectively captures the dynamics of multi-modal data, including aEEG, vital signs, and hemodynamic parameters. Our fusion adaptation of Pathformer demonstrates significant potential in capturing dynamic physiological fluctuations associated with POD.


Through extensive experimentation, we show substantial improvements in sensitivity and diagnostic performance metrics using our Transformer-based framework, highlighting the essential role of advanced representation learning techniques in clinical diagnostics.

\begin{figure}
    \centering
    \includegraphics[width=\linewidth]{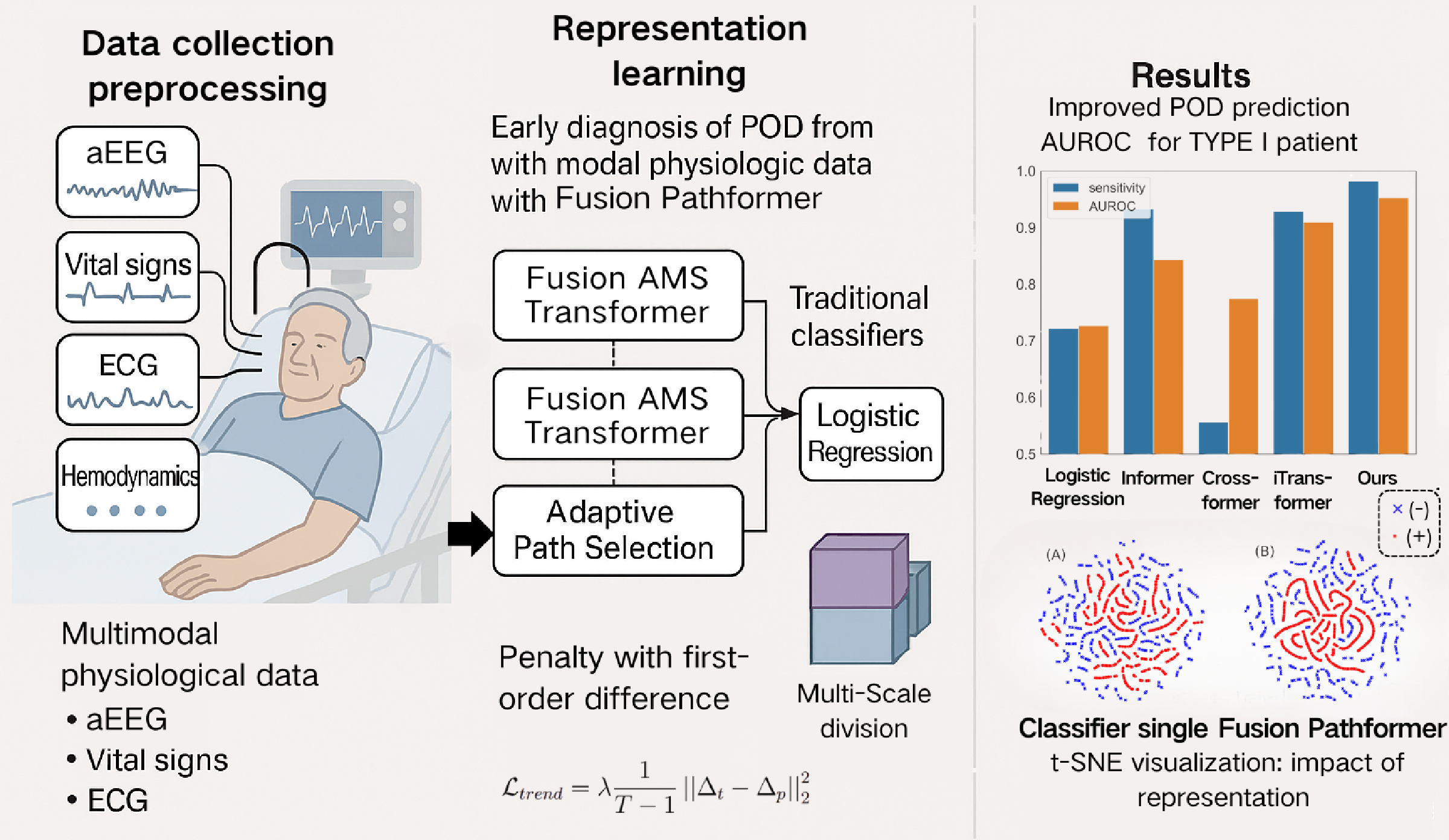}
    \caption{\textbf{Overview of study design.} Post-operative delirium, a prevalent acute neuropsychiatric syndrome is crucial and challenging for early diagnosis. A cascading structure of predictive system was established with components of data preprocessing, Transformer representation learning and machine learning classifier, utilizing multi-modal physiological data including aEEG, vital signs, ECG and hemodynamics. Based on Pathformer\cite{chen2024pathformermultiscaletransformersadaptive}, we proposed fusion adaptation version for multi-modal physiological data and introduced a new regularization item TrendLoss to assist with representation learning. We investigated the average prediction performance of post-operative day 1 to 3 with several Transformer representations. AUROC, sensitivity and representation visualization exhibited a profound improvement with our Fusion Pathformer.}
\end{figure}
\section{Results}\label{sec:res}
\subsection{Patient characteristics and feature selection}
All data were collected from patients admitted to the ICU of The Affiliated Drum Tower Hospital of Nanjing University Medical School following cardiovascular surgery. Physiological data recording commenced postoperatively once the patients were awake and continued for approximately two hours. Clinical diagnoses of POD from postoperative days 1 to 3 were determined by clinicians based on standardized delirium scale scores. Initially, data from 61 patients were collected. After excluding individuals with missing data or incomplete diagnostic records, the final study cohort comprised 56 patients. The surgical procedures included heart valve surgery (18 cases), hemi-aortic arch replacement (17 cases), coronary artery bypass grafting (9 cases), Bentall/Wheat procedures (7 cases), and total aortic arch replacement (5 cases). The average surgical duration was 326.16 minutes. Patients were further classified into two groups (TYPE I and TYPE II) based on the quality of their physiological data. TYPE I included patients whose data mostly remained within acceptable physiological ranges, whereas TYPE II comprised the remaining patients.

To assess the impact of Transformer representation on multi-modal physiological data, we exclusively considered dynamic time-series variables. Static demographic and surgical details, such as age, gender, height, weight, and surgery type, were excluded. Twenty-five physiological characteristics were selected based on Pearson correlation coefficients, ensuring a diverse representation across modalities. Detailed characteristics of both patient groups are summarized in Table \ref{tab:characteristics}.

\newpage\begin{table}[!ht]
\centering
\resizebox{0.8\textwidth}{!}{
\begin{tabular}{lrrrl}
\toprule
Characteristics (Unit) & TYPE I (n=19) & TYPE II (n=37) & P-value \\
\midrule
*$aEEG_{F3}$ & 1.21 (0.19) & 1.16 (0.21) & 0.5151 \\
$aEEG_{P3}$ & 1.02 (0.19) & 0.96 (0.20) & 0.276 \\
*$aEEG_{F4}$ & 1.23 (0.18) & 1.21 (0.21) & 0.9513 \\
$aEEG_{P4}$ & 1.04 (0.17) & 1.01 (0.21) & 0.8598 \\
*CO (L/min) & 4.88 (1.34) & 6.04 (3.11) & 0.0149 \\
CI (L/min/m²) & 2.79 (0.69) & 3.37 (1.67) & 0.0688 \\
*SV (mL) & 62.55 (13.57) & 65.77 (29.98) & 0.4983 \\
SI(mL/$m^2$) & 35.16 (7.23) & 36.64 (15.16) & 0.5795 \\
*SVR ($dyn\cdot s \cdot cm^{\text{-}5}$) & 98.75 (349.61) & 41.23 (46.05) & 0.9444 \\
SVRI($dyn\cdot s \cdot cm^{\text{-}5}/m^2 $) & 18738.69 (54865.58) & 5599.17 (5622.26) & 0.6347 \\
*HR (bpm) & 80.08 (18.44) & 91.56 (23.92) & \textless 0.0001 \\
*MAP (mmHg) & 76.39 (8.61) & 82.04 (13.10) & 0.0023 \\
SYS (mmHg) & 114.41 (11.55) & 118.98 (18.25) & 0.1501 \\
DIA (mmHg) & 60.19 (6.95) & 62.90 (7.53) & 0.0729 \\
*SVV(\%) & 15.28 (6.76) & 19.12 (7.84) & 0.1207 \\
PPV(\%) & 16.48 (6.32) & 23.68 (9.92) & 0.0106 \\
HRV(ms) & 2.65 (1.01) & 4.40 (2.13) & 0.0344 \\
PP(kPa) & 54.22 (13.39) & 56.08 (15.21) & 0.3733 \\
*rSO2\_Ch1(\%) & 71.33 (2.61) & 69.75 (2.39) & 0.7825 \\
*rSO2\_Ch2(\%) & 68.85 (2.21) & 71.68 (3.23) & 0.0077 \\
rSO2\_Ch3(\%) & 34.38 (4.39) & 37.07 (4.74) & 0.5878 \\
rSO2\_Ch4(\%) & 36.62 (5.49) & 42.52 (7.30) & 0.5544 \\
*SpO2 (\%) & 99.71 (0.22) & 99.51 (0.28) & 0.133 \\
*RR (rpm) & 12.37 (1.53) & 13.34 (1.73) & 0.0243 \\
PR (bpm) & 77.91 (10.07) & 88.93 (10.08) & \textless 0.0001 \\
*Temp (\celsius) & 37.41 (0.06) & 37.05 (0.21) & 0.272 \\
*HBP (mmHg) & 115.61 (17.45) & 120.68 (15.18) & 0.0377 \\
LBP (mmHg) & 59.96 (7.90) & 64.41 (9.99) & 0.0166 \\
ABP (mmHg) & 79.05 (7.60) & 83.35 (11.37) & 0.0338 \\
*$Alpha Variability_{F3-P3}$(Hz) & 35.80 (14.09) & 28.71 (12.61) & 0.0044 \\
*$Alpha Variability_{F4-P4}$(Hz) & 34.41 (13.68) & 29.61 (12.37) & 0.0475 \\
*BST(\%) & 14.58 (9.25) & 15.34 (11.33) & 0.1683 \\
3-Delta(\%) & 58.76 (14.87) & 67.53 (14.98) & 0.0015 \\
*3-Theta(\%) & 16.77 (6.95) & 16.63 (7.11) & 0.4675 \\
*3-Alpha(\%) & 14.78 (10.13) & 9.84 (6.26) & 0.0023 \\
*3-Beta(\%) & 9.08 (6.33) & 7.69 (7.33) & 0.027 \\
4-Delta(\%) & 59.66 (15.38) & 65.89 (14.65) & 0.0273 \\
*4-Theta(\%) & 16.13 (6.55) & 17.20 (7.70) & 0.9259 \\
*4-Alpha(\%) & 13.88 (8.97) & 10.34 (6.28) & 0.0332 \\
*4-Beta(\%) & 9.37 (6.35) & 8.28 (7.64) & 0.0551 \\
*Spectral Entropy & 71.64 (5.41) & 68.24 (6.78) & 0.0012 \\
Sex(male) & 14 & 27 & 0.9038 \\
age(y) & 63.63 (7.60) & 62.08 (12.69) & 0.5953 \\
weight(kg) & 70.26 (10.00) & 67.64 (14.24) & 0.5565 \\
height (cm) & 166.32 (5.83) & 165.70 (8.70) & 0.8612 \\
\bottomrule
\end{tabular}}
\caption{\textbf{Patient characteristics and baseline variables(annotated with *).} Data are mean (standard deviation), n.}
\label{tab:characteristics}
\end{table}

\subsection{Performance and visualization of representation}
We assessed the classification performance with and without Transformer representations under identical experimental conditions. The specific Transformer based models evaluated in this study, along with their respective features and methodological differences, are detailed in Table \ref{tab:comparison}.

\begin{table}[!ht] 
      \centering
      \resizebox{\textwidth}{!}{%
      \begin{tabular}{lcccc}
      \hline
      \multicolumn{1}{c}{\textbf{Models}} & 
      \multicolumn{1}{c}{\textbf{Multi-Timestamp}} & 
      \multicolumn{1}{c}{\textbf{Multi-Scale}} & 
      \multicolumn{1}{c}{\textbf{Multi-Channel}} & 
      \multicolumn{1}{c}{\textbf{Token}} \\
      \hline
      Transformer\cite{vaswani2023attentionneed}      &                   &               & $\surd$ & $x_{i,:} \in \mathbb{R}^{d_m}$ \\
      Autoformer\cite{wu2022autoformerdecompositiontransformersautocorrelation}       &                   &               & $\surd$ & $x_{i,:} \in \mathbb{R}^{d_m}$ \\
      Informer\cite{zhou2021informerefficienttransformerlong}         &                   &               & $\surd$ & $x_{i,:} \in \mathbb{R}^{d_m}$ \\
      iTransformer\cite{liu2024itransformerinvertedtransformerseffective}     & $\surd$           &               & $\surd$ & $x_{T,:} \in \mathbb{R}^{T \times d_m}$ \\
      Crossformer\cite{zhang2023crossformer}      & $\surd$           &               & $\surd$ & $x_{i:i+t,j:j+d} \in \mathbb{R}^{t \times d}$ \\
      Origin Pathformer\cite{chen2024pathformermultiscaletransformersadaptive}& $\surd$           & $\surd$       &         & $x_{i:i+t,j} \in \mathbb{R}^{t}$ \\
      Fusion Pathformer (ours) & $\surd$ & $\surd$ & $\surd$ & $x_{i:i+t,j:j+d} \in \mathbb{R}^{t \times d}$ \\
      \hline
      \end{tabular}%
      }
      \caption{\textbf{Token unit comparison between several competitive Transformer architectures.} Existing Transformer models utilize different parts of one time-series $X_{in} \in \mathbb{R}^{T\times d_m}$ as tokens.}
      \label{tab:comparison}
\end{table}

 Logistic Regression(LR) with L2 regularization and Support Vector Machine(SVM) were selected as baseline classifiers. Postoperative delirium(POD) indicators for days 1 to 3 were denoted as PODx, where $x \in [1, 2, 3]$. Performance metrics evaluated included sensitivity, specificity, Youden index, AUROC and AUPRC, calculated as follows:
\begin{align}
    \text{Sensitivity(Recall)} &= \frac{\text{True Positives(TP)}}{\text{TP} + \text{False Negatives(FN)}}\\
    \text{Specificity} &= \frac{\text{True Negatives(TN)}}{\text{TN} + \text{False Positives(FP)}}\\
    \text{Youden index} &= \text{Sensitivity} + \text{Specificity} - 1
\end{align}

Initial experiments were conducted on patient TYPE I to evaluate classification performance with and without Transformer representations. Without representation learning, Logistic Regression and Support Vector Machine classifiers exhibited comparable AUROC values, averaging 0.7274 and 0,7345 respectively. However, when incorporating Transformer representations, both the Origin and Fusion Pathformer models significantly enhanced the AUROC and Youden index, particularly notable for POD2 and POD3 predictions. With Transformer representation, the highest AUROC values for both classifiers exceeded 0.95, demonstrating substantial improvement. All tested Transformer models positively impacted the classification performance. Comprehensive experimental results are detailed in Table \ref{tab:patient1}.

\begin{table}[ht]
    \centering
      \resizebox{\textwidth}{!}{ 
      \begin{tabular}{c|c|cccc|cccc}
          \hline
          \multicolumn{1}{c|}{\multirow{2}{*}{Indicator}} & \multicolumn{1}{c|}{\multirow{2}{*}{Model}} & \multicolumn{4}{c}{Logistic Regression} & \multicolumn{4}{c}{SVM} \\
          \cline{3-10}
          & & \multicolumn{1}{c}{Sensitivity} & \multicolumn{1}{c}{Specificity} & \multicolumn{1}{c}{Youden} & \multicolumn{1}{c|}{AUROC} & \multicolumn{1}{c}{Sensitivity} & \multicolumn{1}{c}{Specificity} & \multicolumn{1}{c}{Youden} & \multicolumn{1}{c}{AUROC} \\
          \hline
          \multirow{6}[2]{*}{POD1} & classifier single & 0.7553 & \textbf{0.7289} & 0.4842 & 0.7409 & 0.7711 & \textbf{0.7737} & \textbf{0.5447} & 0.7468 \\
          & Informer & 0.8000 & 0.4550 & 0.2550 & 0.6840 & 0.8550 & 0.5100 & 0.3650 & 0.7204 \\
          & iTransformer & 0.9350 & 0.6850 & \textbf{0.6200} & 0.8589 & \underline{0.9350} & 0.5100 & 0.4450 & \underline{0.8644} \\
          & Crossformer & 0.5400 & \underline{0.7000} & 0.2400 & 0.7292 & 0.6850 & \underline{0.6550} & 0.3400 & \underline{0.8644} \\
          & Origin Pathformer & \underline{0.9650} & 0.6250 & 0.5900 & \textbf{0.9282} & \textbf{1.0000} & 0.5400 & \underline{0.5400} & \textbf{0.9220} \\
          & Fusion Pathformer(ours) & \textbf{1.0000} & 0.6100 & \underline{0.6100} & \underline{0.9238} & 0.9300 & 0.5500 & 0.4800 & 0.6784 \\
          \hline
          \multirow{6}[2]{*}{POD2} & classifier single & 0.7395 & 0.6263 & 0.3658 & 0.7353 & 0.7263 & 0.6447 & 0.3711 & 0.7485 \\
          & Informer & \textbf{1.0000} & 0.7100 & \underline{0.7100} & 0.8607 & \textbf{1.0000} & 0.4600 & 0.4600 & 0.8319 \\
          & iTransformer & 0.9350 & 0.6200 & 0.5550 & 0.9176 & 0.9350 & 0.4800 & 0.4150 & 0.8599 \\
          & Crossformer & 0.5750 & \textbf{0.7600} & 0.3350 & 0.8046 & 0.5750 & \textbf{0.8050} & 0.3800 & 0.8079 \\
          & Origin Pathformer & \underline{0.9450} & 0.7050 & 0.6500 & \underline{0.9384} & \underline{0.9450} & 0.6500 & \underline{0.5950} & \underline{0.9274} \\
          & Fusion Pathformer(ours) & \textbf{1.0000} & \underline{0.7450} & \textbf{0.7450} & \textbf{0.9905} & 0.9300 & \underline{0.7450} & \textbf{0.6750} & \textbf{0.9340} \\
          \hline
          \multirow{6}[2]{*}{POD3} & classifier single & 0.6711 & 0.5711 & 0.2421 & 0.7061 & 0.6395 & 0.7342 & 0.3737 & 0.7027 \\
          & Informer & \textbf{1.0000} & 0.6400 & 0.6400 & \textbf{0.9859} & \textbf{0.9950} & 0.6400 & 0.6350 & \textbf{0.9659} \\
          & iTransformer & 0.9150 & \underline{0.8450} & \underline{0.7600} & \underline{0.9530} & \underline{0.9750} & 0.6150 & 0.5900 & 0.9240 \\
          & Crossformer & 0.5550 & \textbf{0.8500} & 0.4050 & 0.7901 & 0.6100 & \textbf{0.9150} & 0.5250 & 0.8121 \\
          & Origin Pathformer & 0.9400 & 0.8100 & 0.7500 & 0.9472 & \underline{0.9750} & 0.7350 & \textbf{0.7100} & \underline{0.9451} \\
          & Fusion Pathformer(ours) & \underline{0.9450} & 0.8200 & \textbf{0.7650} & 0.9438 & 0.9300 & \underline{0.7500} & \underline{0.6800} & 0.9304 \\
          \hline
      \end{tabular}
      }
      \caption{\textbf{The classification results of baseline classifiers with and without 5 Transformer representations.} Prediction period $T = 30 \text{ minutes}$ and TrendLoss regularization strength $\lambda = 5^{-4}$. The best results are highlighted in bold across different models, and the second-best results are underlined.}
      \label{tab:patient1_in}
\end{table}

To validate improvements in classification results achieved by representations, we conducted dimensionality reduction using t-distributed Stochastic Neighbor Embedding (t-SNE). We visualized data distributions before and after applying Transformer representations, facilitating a clear demonstration of the improved separation between positive (POD) and negative (non-POD) patient samples.

\begin{figure}[ht]
    \centering
    \includegraphics[width=\linewidth]{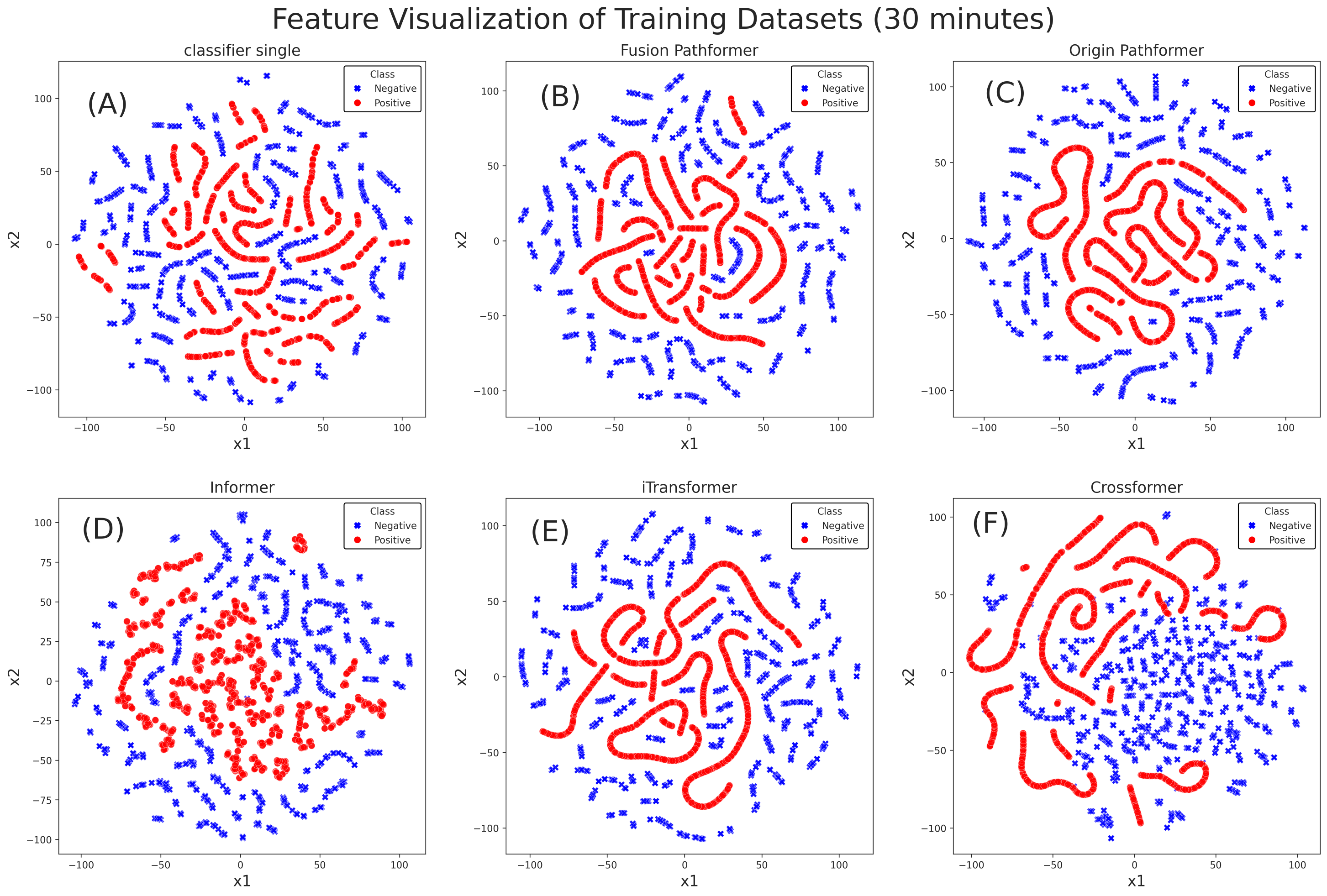}
    \caption{\textbf{Visualization of training set using t-SNE dimension reduction.} Positive samples came from patients diagnosed POD on the first post-operative day while negative samples came from those diagnosed normal. We generated the samples by window sliding, the total sample size is 1824 with prediction period of 30 minutes.}
    \label{fig:visualization}
\end{figure}

Visualization with t-SNE dimensionality reduction (Figure \ref{fig:visualization}) reveals distinct differences in data distributions before and after Transformer representation. Initially(Figure \ref{fig:visualization}(A)), positive samples are largely concentrated in the central region, with negative samples predominantly dispersed toward the periphery, although with some overlap. After applying the Fusion Pathformer representation, positive samples display increased clustering and clearer delineation from negative samples, despite some residual overlap. Conversely, Crossformer representation results in notably different data distribution, positioning positive and negative samples towards opposite ends of the visualization space. This distinct separation suggests that Crossformer's cross-dimensional attention mechanism may effectively discriminate between patient subtypes, a hypothesis warranting further detailed investigation. This may be a reason that representation by Crossformer can achieve the best among all the Transformer architectures in all POD indicators.

\subsection{Ablation and hyperparameter settings}
To further validate the effectiveness of our proposed Fusion Pathformer and TrendLoss methods, ablation studies were conducted without altering other hyperparameters. The performance metrics, including sensitivity, specificity, and AUROC, are summarized in Table 4. Across all POD indicators, Fusion Pathformer combined with TrendLoss consistently outperformed other models, achieving the highest Youden index and AUPRC. Specifically, the Youden indices for POD1, POD2, and POD3 were 0.6100, 0.7450, and 0.7650, respectively. These findings highlight the robust performance enhancement provided by our fusion adaptation and underscore the beneficial impact of incorporating TrendLoss regularization.

 However, this improvement is classifier-related. Origin Pathformer without TrendLoss seemed to be more competitive with SVM. 

\begin{table}[ht] 
      \centering
      \resizebox{\textwidth}{!}{  
      \begin{tabular}{c|c|ccc|ccc}
            \hline
            \multicolumn{1}{c|}{\multirow{2}{*}{Indicator}} & \multicolumn{1}{c|}{\multirow{2}{*}{Model}} & \multicolumn{3}{c}{Logistic Regression} & \multicolumn{3}{c}{SVM} \\
            \cline{3-8}
            & & \multicolumn{1}{c}{Sensitivity} & \multicolumn{1}{c}{Specificity} & \multicolumn{1}{c|}{AUROC} & \multicolumn{1}{c}{Sensitivity} & \multicolumn{1}{c}{Specificity} & \multicolumn{1}{c}{AUROC} \\
            \hline
            \multirow{5}[2]{*}{POD1} & classifier single & 0.7553 & \textbf{0.7289} & 0.7409 & 0.7711 & \textbf{0.7737} & 0.7468 \\
            & Origin without TrendLoss & 0.9550 & \underline{0.6300} & \underline{0.9276} & \underline{0.9750} & \underline{0.6250} & \underline{0.8869} \\
            & Origin with TrendLoss & \underline{0.9650} & 0.6250 & \textbf{0.9282} & \textbf{1.0000} & 0.5400 & \textbf{0.9220} \\
            & Fusion without TrendLoss & 0.9500 & 0.5650 & 0.8865 & 0.9250 & 0.5100 & 0.6993 \\
            & Fusion with TrendLoss & \textbf{1.0000} & 0.6100 & 0.9238 & 0.9300 & 0.5500 & 0.6784 \\
            \hline
            \multirow{5}[2]{*}{POD2} & classifier single & 0.7395 & 0.6263 & 0.7353 & 0.7263 & 0.6447 & 0.7485 \\
            & Origin without TrendLoss & 0.9550 & 0.6550 & 0.9337 & \underline{0.9350} & \underline{0.7150} & 0.9178 \\
            & Origin with TrendLoss & 0.9450 & \underline{0.7050} & 0.9384 & \textbf{0.9450} & 0.6500 & 0.9274 \\
            & Fusion without TrendLoss & \underline{0.9700} & 0.6650 & \underline{0.9718} & 0.9300 & \textbf{0.7450} & \textbf{0.9342} \\
            & Fusion with TrendLoss & \textbf{1.0000} & \textbf{0.7450} & \textbf{0.9905} & 0.9300 & \textbf{0.7450} & \underline{0.9340} \\
            \hline
            \multirow{5}[2]{*}{POD3} & classifier single & 0.6711 & 0.5711 & 0.7061 & 0.6395 & 0.7342 & 0.7027 \\
            & Origin without TrendLoss & 0.9350 & 0.7750 & 0.9351 & \textbf{0.9750} & \textbf{0.8050} & \underline{0.9593} \\
            & Origin with TrendLoss & 0.9400 & \underline{0.8100} & \underline{0.9472} & \textbf{0.9750} & 0.7350 & 0.9451 \\
            & Fusion without TrendLoss & \textbf{0.9700} & 0.7600 & \textbf{0.9531} & \underline{0.9600} & \underline{0.7800} & \textbf{0.9700} \\
            & Fusion with TrendLoss & \underline{0.9450} & \textbf{0.8200} & 0.9438 & 0.9300 & 0.7500 & 0.9304 \\
            \hline
      \end{tabular}
      }
      \caption{\textbf{Ablation study conducted on fusion adaptation and TrendLoss.} We only use Pathformer as an example with prediction period $T = 30 \text{ minutes}$ and TrendLoss regularization strength $\lambda = 5^{-4}$. The best results are highlighted in bold across different models, and the second-best results are underlined.}
      \label{tab:ablation_in}
\end{table}

\begin{figure}[ht]
    \centering
    \includegraphics[width=\textwidth]{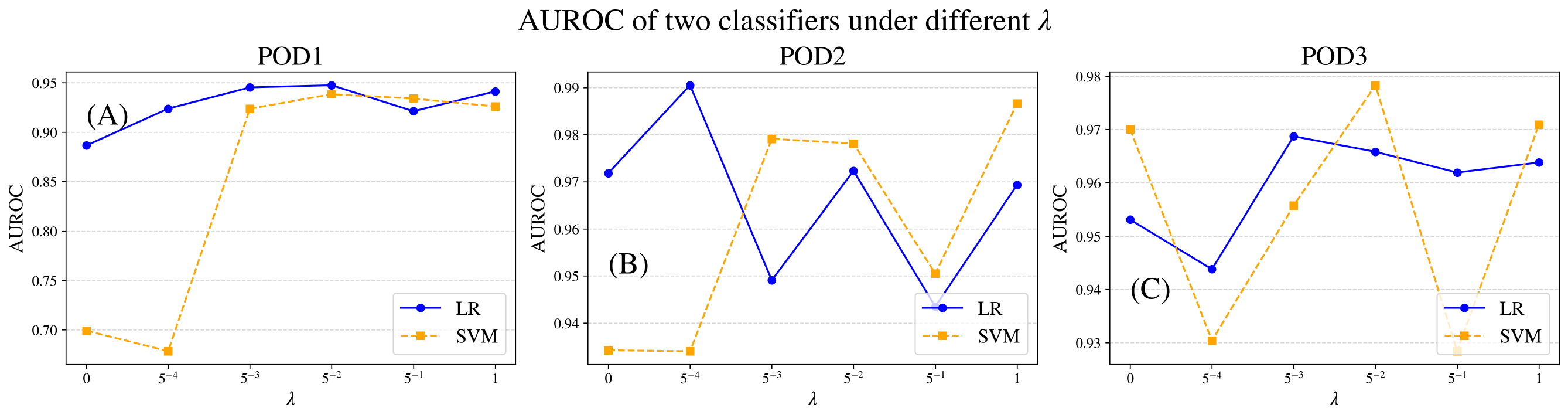}
    \caption{\textbf{Impact of TrendLoss regularization strength $\lambda$.} Different results on POD1-3 emphasizes different optimal choices for different POD indicators and different machine learning classifiers.}
    \label{fig:trend_loss}
\end{figure}

In addition to the novel methodologies introduced, we investigated key hyperparameters influencing Transformer representation performance, specifically the regularization strength $\lambda$ of TrendLoss and the prediction period $T$. As demonstrated in Figure \ref{fig:trend_loss}, optimal classification outcomes were achieved with varying $\lambda$ across POD indicators (POD1-3), highlighting the need for careful tuning this hyperparameter.

\begin{figure}[ht]
    \centering
    \includegraphics[width=\linewidth]{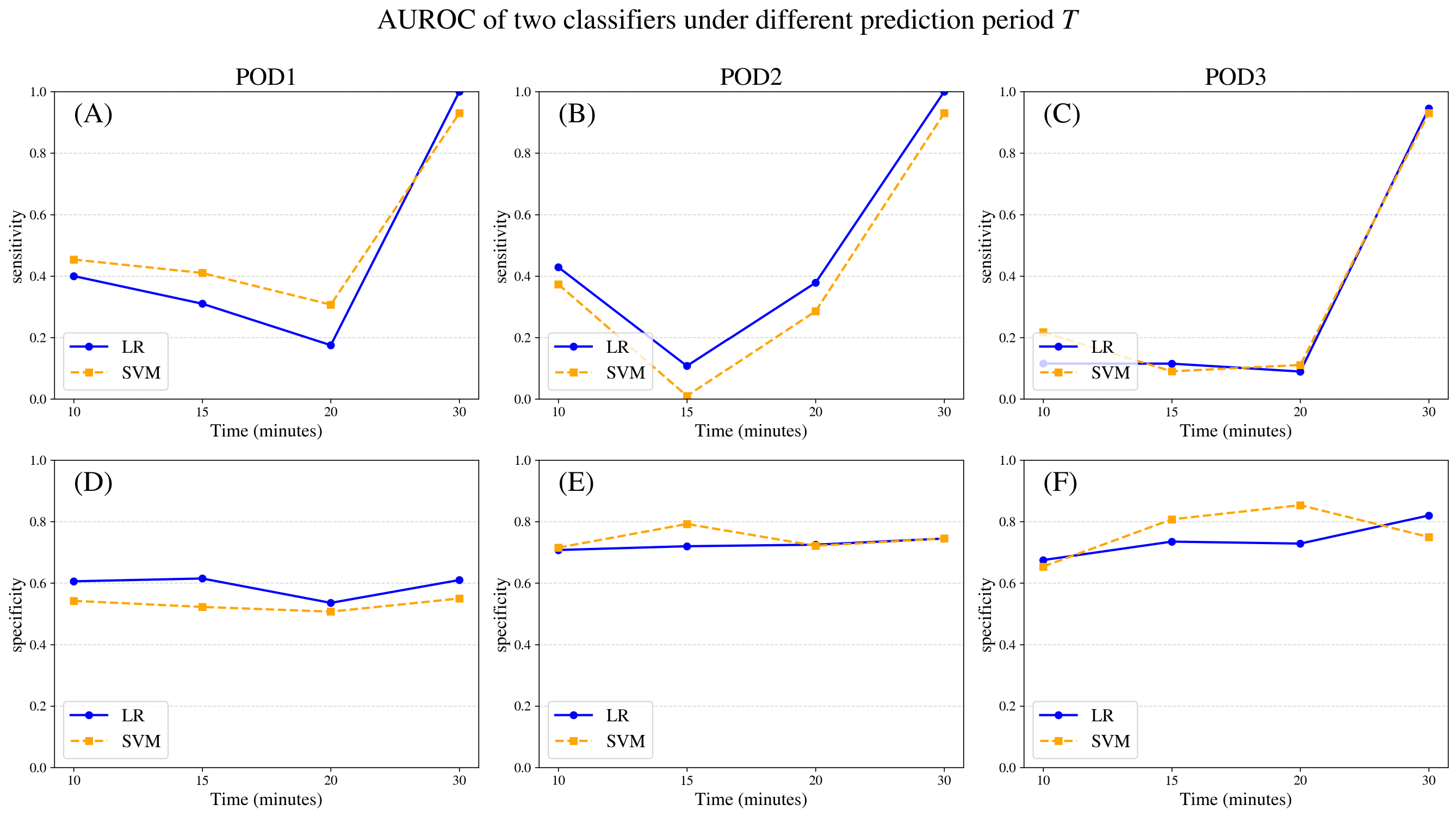}
    \caption{\textbf{Impact of varying prediction period $T$.} Results of Logistic Regression and SVM under all POD indicators exhibit the similar pattern when prediction period $T$ is changing. Subfigure (A)(B)(C) indicate that the diagnosis of POD patients is quite sensitive to the prediction period $T$. These results emphasize that macrospoic information and longer monitoring duration are critical for POD diagnosis.}
    \label{fig:prediction_period}
\end{figure}

To investigate the influence of prediction period $T$ during the representation learning phase, we varied the temporal resolution while maintaining a fixed number of timestamps per time series. Specifically, by sampling 180 records at intervals of 5 seconds, we obtained a 15-minute duration, whereas sampling at intervals of 10 seconds yielded a 30-minute period. This approach allowed us to differentiate between macroscopic (coarse-grained) and microscopic (fine-grained) physiological information. As illustrated in Figure \ref{fig:prediction_period}, longer prediction periods significantly enhanced model performance, suggesting that capturing macroscopic physiological fluctuations is essential for effective representation learning and subsequent POD prediction.

\subsection{Distinct outcomes on patient TYPE II}
Given the distinct preprocessing methods applied to different patient subtypes, we assessed the efficacy of various Transformer representations under standardized settings. Transformer architectures evaluated for patient TYPE II included Transformer\cite{vaswani2023attentionneed}, Autoformer\cite{wu2022autoformerdecompositiontransformersautocorrelation}, Informer\cite{zhou2021informerefficienttransformerlong}, Crossformer\cite{zhang2023crossformer}, iTransformer\cite{liu2024itransformerinvertedtransformerseffective}, and Pathformer\cite{chen2024pathformermultiscaletransformersadaptive}, with detailed outcomes presented in Table \ref{tab:patient2} of the appendix. Notably, Transformer and Autoformer demonstrated significant performance improvements, particularly evident in AUROC and AUPRC across all POD indicators. For instance, the Transformer model achieved an AUROC of 0.8622 for POD3, a notable increase compared to the baseline SVM classifier. Informer similarly showed robust performance for patient TYPE II. Notably, these three Transformer models process each dimension independently, applying self-attention exclusively in the temporal dimension. In contrast, models employing patch mechanisms demonstrated increased specificity but markedly decreased sensitivity, with Fusion Pathformer yielding the lowest sensitivity. These findings suggest temporal dependencies may play a more crucial role than inter-dimensional relationships in distinguishing POD from non-POD patients in TYPE II.

\section{Discussion}
Postoperative delirium (POD) is an acute disturbance of brain function associated with increased morbidity and mortality\cite{RN6,10.1001/jama.2017.12067}. Its complex pathophysiology involves multiple factors such as neuroinflammation, cerebral hypoperfusion, and systemic stressors, particularly evident in cardiovascular surgical patients\cite{10.1001/jamanetworkopen.2023.37239, Raats2015-om}. Here, we investigated the potential of multi-modal physiological data representation using advanced Transformer architectures, focusing specifically on a clinical adaptation through our proposed Fusion Pathformer.

Our findings demonstrate that Transformer models effectively capture the temporal dynamics inherent in multi-modal physiological signals, significantly enhancing predictive performance for POD. Particularly, our Fusion Pathformer adaptation underscores the value of multi-modal Transformer architectures in clinical scenarios. These results align with previous work, such as Medformer\cite{wang2024medformermultigranularitypatchingtransformer}, emphasizing Transformers' capabilities not merely as forecasting tools but as robust methods for representation learning in medical time-series data.

Through extensive experimental evaluations, we have provided key insights highlighting the necessity of advanced Transformer-based methodologies for improving POD diagnostic accuracy and facilitating better clinical decision-making.

Transformer representations enhance sensitivity for POD1-3 indicators but can adversely affect specificity, particularly for POD1. As shown in Table \ref{tab:patient1_in}, the Fusion Pathformer and iTransformer consistently improved sensitivity, while Crossformer notably improved specificity. This suggests that Crossformer's attention mechanism, which integrates relationships across multiple dimensions, may preferentially distinguish non-POD patients by emphasizing both inter-dimensional and temporal relationships. 

Among patch-based Transformer models, the Fusion Pathformer notably demonstrated the greatest sensitivity enhancement. Fusion/Origin PathFormer, Crossformer are three Transformers that implement patch mechanism. In the Table \ref{tab:patient1_in}, our Fusion Pathformer significantly improved sensitivity, likely due to its enhanced capacity to  effectively capture the pronounced abnormal physiological fluctuations observed in POD patients with its adaptive multi-scale router. In constrast, the fixed length segmentation in Crossformer may have obscured the critical temporal features. 

Temporal features appear particularly crucial in identifying POD patients. Informer, which focuses exclusively on the attention over the whole time-series, often surpassed Crossformer, even sometimes fusion Pathformer in sensitivity. This may imply that temporal characteristics are essential than relationship between different dimensions in POD patients. This appears intuitive, as POD patients often exhibit noticeable changes in their physiological signals like the drastic fluctuation of heart rate(HR) as well as a slowdown in electroencephalographic activity. This suggests that in clinical monitoring, continuously tracking variations in a single-modality even one physiological signal may be more beneficial.

Transformer representations show preference for Logistic Regression compared to Support Vector Machine. According to Table \ref{tab:patient1_in}, improvements in the Youden index were substantially greater for LR than for SVM following Transformer representations. These findings indicate that Transformer-derived features align better with linear classifiers such as Logistic Regression for POD prediction tasks.

TrendLoss enhances downstream classification performance, particularly when pre-training involves high prediction error. Comparing Fusion Pathformer trained with and without TrendLoss, the former consistently exhibited superior sensitivity, specificity and AUROC metrics when coupled with LR. Given the complexity of high-dimensional, long-term physiological data, reliance solely on mean squared error(MSE) appears insufficient, suggesting the necessity of additional regularization methods such as TrendLoss. 

Macroscopic and coarse-grain temporal information plays crucial role in POD diagnosis. Experiments utilizing varying prediction period revealed substantial performance differences. Specifically, classification results significantly deteriorated when the prediction period was shorter than 30 minutes, emphasizing that macroscopic information is pivotal for POD diagnosis as they usually provide more information about the fluctuation and changes in the physiological state of one patients. Consequently, monitoring periods of at least 30 minutes are recommended for reliable clinical delirium assessments, aligned with another research using Transformer to predict POD on intraoperative temporal dynamics\cite{Giesa2024-mu}. 

Logistic Regression and SVM alone fail in effectively distinguish between POD and non-POD patients for TYPE II. Experiments conducted on patient TYPE II reflect that standalone LR and SVM classifiers are incapable of differentiating POD patients from non-POD patients, according to Table \ref{tab:patient2}. The extremely low sensitivity suggests that distinctions between POD and non-POD patients are not well manifested in the raw physiological signals. This highlights the necessity of advanced representation methods to capture informative patterns in such challenging patient cohorts.

Transformer architectures focused exclusively on temporal dependency outperform other Transformer variants for patient TYPE II. Our results show that Transformer models like Transformer, Autoformer and Informer, which apply self-attention solely in the temporal dimension and involve no information about different modalities, substantially improved both sensitivity and specificity, particularly evident for POD3. These findings suggest that for patient TYPE II, temporal fluctuations are more clinically informative than inter-dimensional relationships, although specificity enhancements observed in other Transformer variants are also noteworthy.

Recent researches on postoperative delirium (POD) increasingly explore predictive methodologies leveraging both traditional machine learning and deep learning techniques. Established machine learning approaches such as Logistic Regression, Support Vector Machines (SVM), Random Forest, and XGBoost have shown effectiveness\cite{Lee2023, Nagata2023}, predominantly relying on static pre- or post-operative patient data. In contrast, deep learning methods have typically utilized relatively simple neural network structures or have directly classified single-modal data using complex architectures like Long Short-Term Memory(LSTM) and Transformers\cite{Giesa2024-mu, Mulkey2023}. This indicates a clear gap in the comprehensive application of sophisticated representation learning methods for multi-modal physiological data in POD diagnosis.

Our proposed framework specifically addresses this gap by introducing a Transformer-based representation learning model tailored for dynamic, multi-modal physiological signals. Nevertheless, our study is constrained by limitations, including a relatively small patient cohort and potential biases stemming from patient selection. Furthermore, variability in anesthetic protocols and intraoperative management strategies may introduce additional confounding factors, affecting data consistency and comparability. In data preprocessing, we recognize the potential benefits of adopting more advanced and efficient approaches. Currently, due to scalability concerns, our framework employs simple linear interpolation methods rather than more sophisticated deep learning techniques for handling missing data.

Additionally, we advocate for more researchers to contribute to the development of comprehensive public datasets that include diverse modalities and a large cohort of patients. Such resources will significantly enhance predictive accuracy and diagnostic capabilities for postoperative delirium(POD) and related conditions.

To summarize, our study introduced a cascarding framework utilizing Transformer representation learning tailored for multi-modal physiological data. The proposed Fusion Pathformer achieved a substantial improvement of over 25\% compared to baseline Logistic Regression and Support Vector Machine classifiers for POD prediction of postoperative day 1 to 3. Through detailed analysis and eight 
 key observations, our findings demonstrate that Transformer-based representation significantly improves patient stratification, particularly benefiting elderly individuals undergoing cardiovascular surgery.

\section{Methods}

\subsection{Data collection and preprocessing}
We collected five modalities of data from cardiothoracic ICU, including \textbf{basic information}(height, weight, etc.), \textbf{vital signs}(body temperature, oxygen saturation, etc.), \textbf{electrocardiographic monitor recordings}(mean arterial pressure, cardiac output, etc.), \textbf{regional cerebral oxygenation signals} as well as selected amplitude-integrated electroencephalography (aEEG) parameters. Detailed information is shown in Table \ref{tab:characteristics}. All the data we considered in the study is time-series physiological signals but with variable resolutions. 

Prior to exploring the features of multi-modal data, essential preprocessing steps were conducted as illustrated in Figure \ref{fig:framework}. Initially, we screened the collected data to retain patients with complete recordings and full annotated POD indicators. Ultimately, data from 56 patients were selected. Preliminary visualization revealed significant individual variability and identified noise disturbances caused by environmental factors, medical instruments, and patients movements. Consequently, we implemented a sanity check to filter out suspicious recordings. Patients with over 10\% anomalous data were classifierd as TYPE II. Patients whose data predominantly fell within acceptable range shown in Figure \ref{fig:framework} were designated as TYPE I. 

Subsequent to this sanity check, we applied one more step of anomaly repair to fix the outliers specifically to patient TYPE I. We calculated the dynamic mean of each feature and replaced the suspicious recordings. Missing values in both types were filled using linear interpolation\cite{Morelli2019-rw}. Given the varying recording frequencies and inconsistent recording intervals across modalities, temporal alignment was crucial for integration. Thus, we standardized the temporal resolution to a 1-second baseline, applying padding where necessary to maintain uniformity across modalities.

To simplify the design and reduce the computational load of Transformer representation model, we applied downsampling and exponential smoothing techniques to prepare the training and testing data for downstream task. 

\subsection{Problem definition}
Currently, limited research has examined the potential of multi-modal physiological representation learning in disease diagnosis, particularly regarding the early detection of postoperative delirium(POD). To establish a standardized framework beneficial to future research, we adopted and extended the definition of Medical Time Series Classification proposed in MedFormer\cite{wang2024medformermultigranularitypatchingtransformer} to suit the context of our study.

\begin{problemx}
Given the multi-modality physiological data of one patient undergoing cardiovascular surgery $X_{in}\in \mathbb{R}^{T\times D}$, where $T$ is the length of time-series and $D = \sum_{1\leq m\leq M}d_m$ is the characteristics dimensions. We aim to learn a uniform representation of the data $\mathbf{h}$ which can be used to predict the PODx where $x\in \{1, 2, 3\}$ indicating whether he/she will slipping into postoperative delirium in the next 3 days.
\end{problemx}

Here, a uniform representation is learned by increasingly powerful and widely used time-series representation tool, Transformer architecture.
\subsection{Framework building and module design}\label{sec:framework}
\begin{figure}[!ht]
    \centering
    \includegraphics[width=0.9\textwidth]{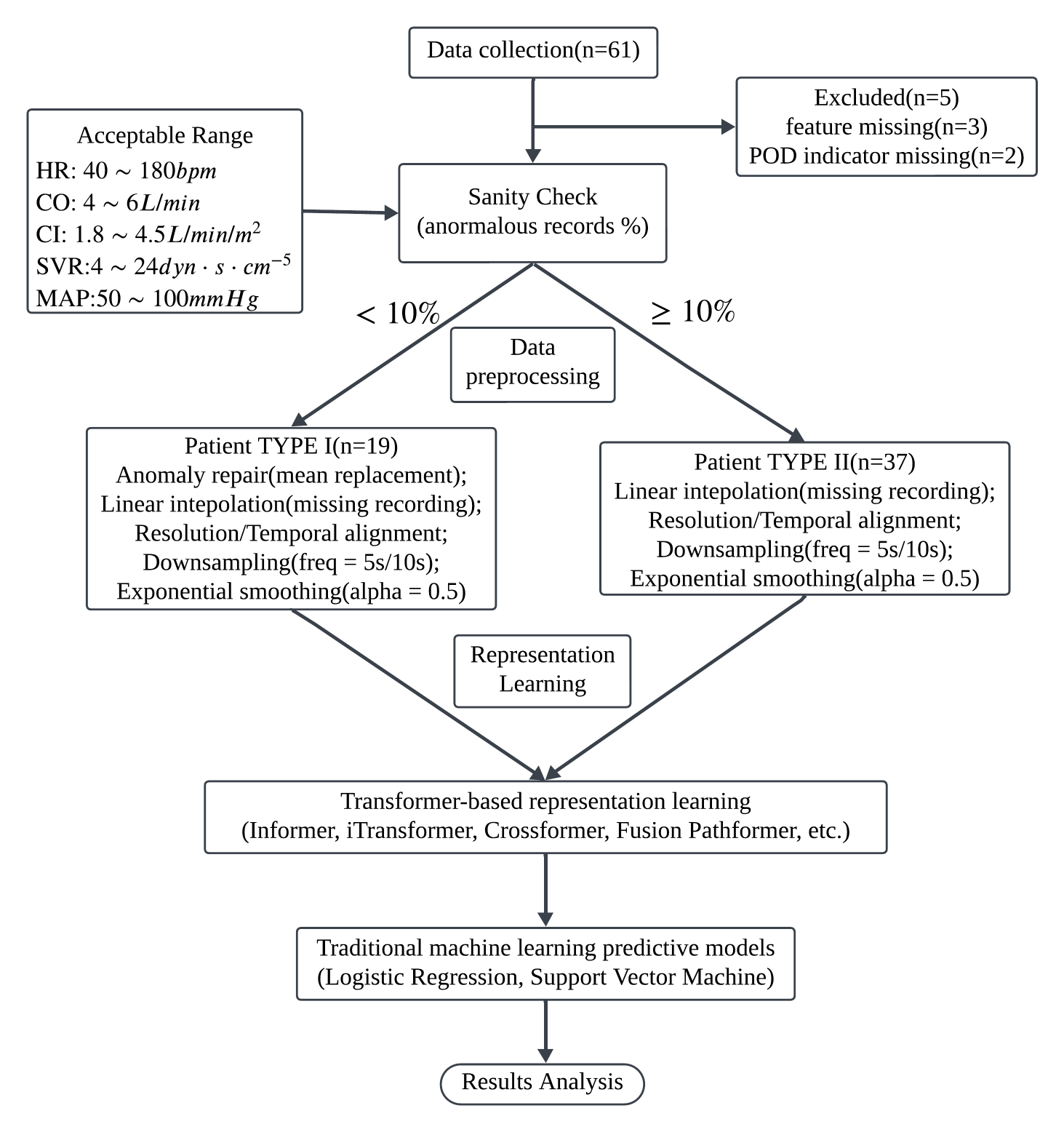}
    \caption{\textbf{Our design of representation-based post-operative delirium prediction model.} The model comprises three main parts, data preprocessing, representation learning and predictive classification learning.}
    \label{fig:framework}
\end{figure}

As in Figure \ref{fig:framework}, we developed a cascading structure starting from data collection and concluding at result analysis. Following preprocessing, the data were standardized and input into Transformer representation learning model. Subsequently, we employed traditional machine learning classifiers like Support Vector Machine(SVM) as well as Logistic Regression(LR) to accomplish the prediction task.

Transformer models have merged as prominent tools for time-series prediction, demonstrating exceptional performance across diverse domains such as natural language processing(NLP)\cite{brown2020languagemodelsfewshotlearners, raffel2023exploringlimitstransferlearning} and Computer Vision(CV)\cite{dosovitskiy2021imageworth16x16words, liu2021swintransformerhierarchicalvision, carion2020endtoendobjectdetectiontransformers}. Their success in NLP and CV has encouraged recent researches on Transformer adaptation to medical domain. Recent studies, including PT3\cite{10191261} and MedFormer\cite{wang2024medformermultigranularitypatchingtransformer}, have leveraged Transformer architectures to address clinical tasks like sepsis risk prediction and multi-granularity health pattern analysis. Specifically, PT3 introduced a dual-task pretraining framework utilizing classical Transformer models to predict mortality risks in sepsis, whereas MedFormer applied multi-granularity patching methods to represent medical time series data effectively. Building upon these advances, our study explores the capability of a novel Transformer representation model, Fusion Pathformer, tailored specifically for multimodal physiological signals in predicting postoperative delirium (POD).

Meantime, we proposed a fusion adaptation of Pathformer\cite{chen2024pathformermultiscaletransformersadaptive} as a new backbone. Our adaptation can be summarized as: different modalities share the same transformer-like block, namely AMS block proposed in original Pathformer to share the knowledge between each other. Router component will capture the dynamic characteristic of each modality and allocate proper Transformer blocks for each of them. This means different modalities may be assigned to different AMS blocks according to their dynamic features and utilize \textbf{Patching mechanism}\cite{nie2023timeseriesworth64, xue2024cardchannelalignedrobust} to partition time-series into parts. 

\subsubsection{Multi-modal adaptation of Pathformer}
\begin{figure}[ht]
    \centering
    \includegraphics[width=\linewidth]{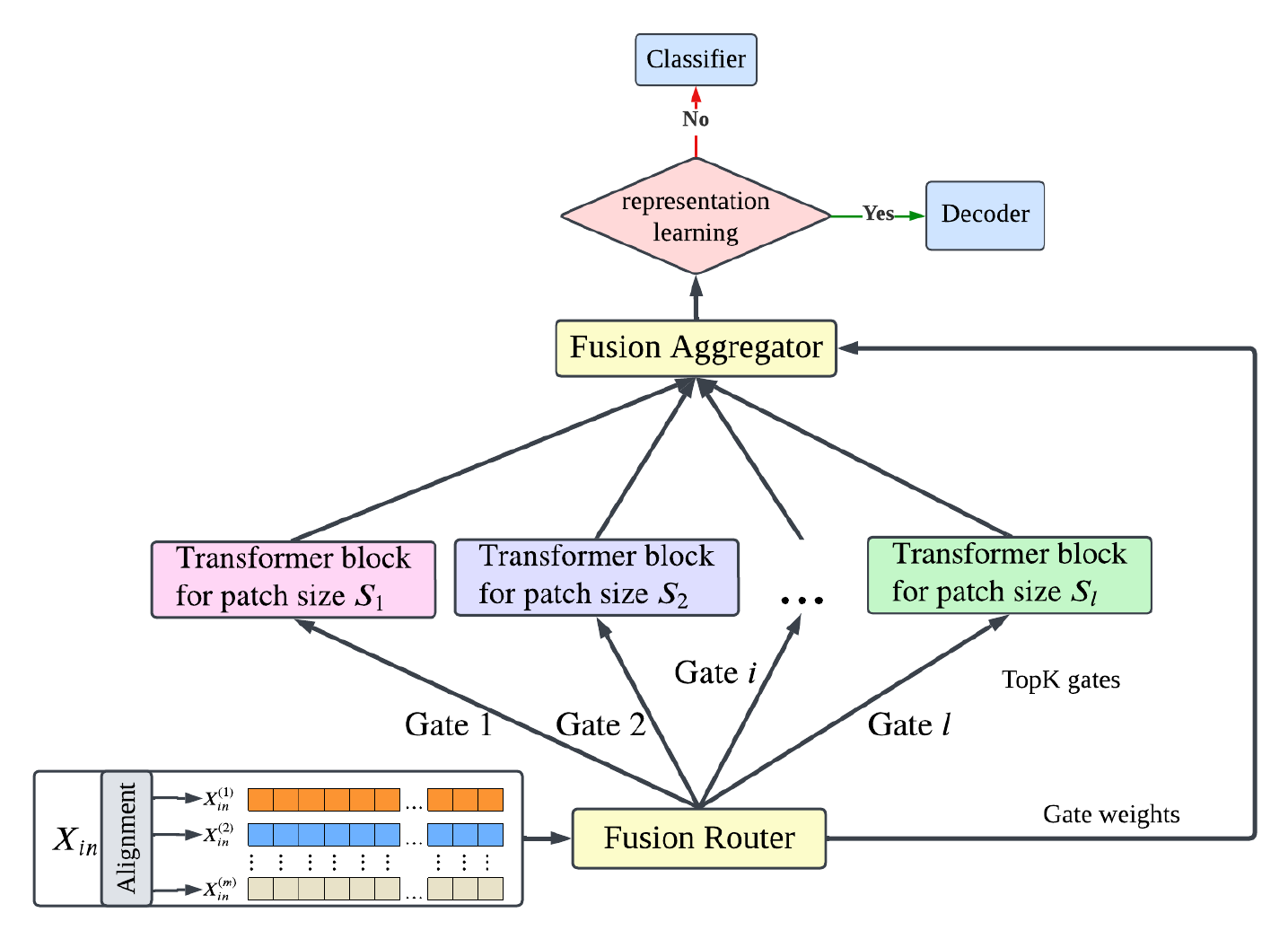}
    \caption{\textbf{Fusion adaptation of Pathformer.} Each modality in one input time-series will be routed to topK gates each of which is connected with one Transformer block. Before inputting each modality into Transformer blocks, alignment is done to embed different modalities into the same size. Each small square in the figure corresponds to one timestamp $X_{i}^{(m)} \in \mathbb{R}^{d_e}$.}
    \label{fig:fusion}
\end{figure}

Implementing individual Transformer blocks for each modality and patch size would be computationally infeasible. Therefore, we integrated information from different modalities into a unified representation framework, significantly reducing computational complexity and time consumption (Figure \ref{fig:fusion}). Our proposed Fusion Pathformer leverages fundamental design principles of Pathformer, including multi-scale segmentation and dual attention mechanisms, optimizing it specifically for multi-modal physiological data.

Focusing initially on a single modality, denoted as $X_{in}^{(m)} \in \mathbb{R}^{T\times d_m}$, where $d_m$ denotes the dimension of this modality, we then integrated them together to get the final unified representation. Given the variability of dimensions across modalities, we applied a learnable alignment embedding.

\begin{equation}
    X_{in} = X_{in}^{(m)} W_{a}^{(m)} + b_{a}^{(m)}
\end{equation}
where $W_{a}^{(m)} \in \mathbb{R}^{d_m \times d_e}$ and $b_{a}^{m} \in \mathbb{R}^{d_e}$.
Subsequently, we adapted and reformulated the equations specifically for application within the multi-modal context.

\begin{itemize}
    \item \textbf{Multi-scale division}: We provide multi-scale mechanism for each modality using a collection of candidate patch sizes $\mathrm{L} = \{S_1, \cdots, S_L\}$. Each $S_i$ corresponds to a distinct patch division operation. An input time-series of length $T$ will be segmented into $P = T/S_i$ temporal patches $X_{in}^{(p)} \in \mathbb{R}^{S_i\times d_e}$. 
    \item \textbf{Dual attention and Multi-scale Router}: Following the fundamental Pathformer design, for attention calculation, we implemented a channel-dependent embedding, distinguishing our method from Origin Pathformer. Additionally, the original multi-scale router implementation was retained for its demonstrated effectiveness.
    \item \textbf{Fusion Aggregator}: After multi-scale router, each modality $X_{in} \in\mathbb{R}^{T\times d_e}$ will choose the top K pathway adaptively, we group them by their choices. This means all the modalities choosing the same pathway will share the transformer block followed by. We denote $G_{S_l}$ as the set consisting of all time series that choose the patch size $S_l$ and $Tr_{l}$ as the fusion transformer block for patch size $S_l$. 

    The output of Fusion aggregator is a little different from the original one.

    \begin{equation}
        X_{out} = \sum_{l=1}^{L} I(\bar{R}(X_{trans})_l > 0) R(X_{trans})_l Tr_l(X_{in})
    \end{equation}
    where $R(\cdot)$ is routing function which will route time series to its matched gate $G_{S_l}$ and assign a weight to this gate. $I(\cdot)$ is an indicator. $X_{out} \in \mathbb{R}^{T\times d_e}$ for each modality.

    To fuse the information between different modalities, we stacked all the $X_{out}$ to get the representation.

    \begin{equation}
        \bar{X}_{in} = \text{Stack}(X_{out}^{(1)}, \cdots, X_{out}^{(M)}) \in \mathbb{R}^{T\times M \times d_e}
    \end{equation}

    To get the uniform representation we state in the problem definition, a weight vector $W_h\in \mathbb{R}^{M}$ and temporal average will be applied to $\bar{X}_{in}$.
    
    \begin{align}
        X_{h} &= \text{MatMul}(\bar{X}_{in}, W_h) \in \mathbb{R}^{T\times d_e} \\
        h&=\text{Average}_t(X_{h}) \in \mathbb{R}^{d_e}
    \end{align}
\end{itemize}

\subsection{Training strategy}\label{sec:train}
We constructed the full framework in a triple-stage manner where the last two stages involved model training (Figure \ref{fig:framework}). The training and testing approaches for these two stages are quite different.

\subsubsection{Transformer representation learning and inference}
Auto-regression and self-supervision are two frequently encountered tasks and methodologies in time series analysis\cite{Senane_2024, foumani2023series2vecsimilaritybasedselfsupervisedrepresentation}. Auto-regression, also prestigious for time series forecasting, has been extensively studied\cite{oord2019representationlearningcontrastivepredictive}. In recent years, researches based on the Transformer architecture has proliferated, like Autoformer\cite{wu2022autoformerdecompositiontransformersautocorrelation}, Informer\cite{zhou2021informerefficienttransformerlong}, Crossformer\cite{zhang2023crossformer}, iTransformer\cite{liu2024itransformerinvertedtransformerseffective}, all concentrating on time series prediction with auto-regression. 

We selected auto-regression task for representation learning. To accomplish this goal, we add a linear decoder to project the $\bar{X}_h$ to $X_{pred}$. 

\begin{equation}
    X_{pred} = \text{Linear}(X_{h}) \in \mathbb{R}^{T\times D}
\end{equation}

\textbf{TrendLoss}\quad Tendency is a crucial feature in time-series data. Therefore, we incorporate the differences in sequence trend variations as a penalty term, analogous to regularization methods, as part of the training loss.

\begin{definition}
Given time series target $X_{target} \in \mathbb{R}^{T\times d}$ and model prediction $X_{pred} \in\mathbb{R}^{T\times d}$, we denote $\Delta_{t}, \Delta_{p} \in \mathbb{R}^{(T-1) \times d}$ as the first-order difference of $X_{target}$ and $X_{pred}$ respectively. Trend Penalty is:
\begin{equation}
\mathcal{L}_{trend} = \lambda \frac{1}{T-1}\left|\left| \Delta_{t} - \Delta_{p}\right|\right|_2^2
\end{equation}
where $\left|\left| \cdot \right|\right|_p$ is the p-norm.
\end{definition}
\subsubsection{Classifier training and testing}
In classifier training and testing phase, to evaluate the effect of Transformer representation, we designed two different dataset preparation methods to compare the differences under identical settings with and without Transformer representations.

\begin{align}
\begin{cases}
    X = \text{Average}_t(X_{in})\in\mathbb{R}^{d_m} & \text{without representation}\\
    X = h \in \mathbb{R}^{d_e}&\text{with representation}
\end{cases}
\end{align}

Class imbalance (ratio of noncases to cases > 1) routinely occurs in clinical scenarios and may degrade the predictive performance of machine learning algorithms\cite{Cartus2020-pr}. After dataset preparation, random undersampling is applied to balance the categories of samples prior to inputting the data to the models, which is also a usual approach used in conventional clinical machine learning algorithms researches\cite{kanbar2018undersamplingbaggingdecisiontrees}.

As we worked on small cohort of patients, we adopted subject-dependent strategy instead of subject-independent strategy\cite{wang2024medformermultigranularitypatchingtransformer}. 

\bibliography{manuscript}
\newpage
\section{Appendix}
\subsection{Experimental results}
\begin{table}[!ht] 
      \centering
      \resizebox{\textwidth}{!}{  
      \begin{tabular}{c|c|ccccc|ccccc}
          \hline
          \multicolumn{1}{c|}{\multirow{2}{*}{Indicator}} & \multicolumn{1}{c|}{\multirow{2}{*}{Model}} & \multicolumn{5}{c}{Logistic Regression} & \multicolumn{5}{c}{SVM} \\
          \cline{3-12}
          & & \multicolumn{1}{c}{Sensitivity} & \multicolumn{1}{c}{Specificity} & \multicolumn{1}{c}{Youden} & \multicolumn{1}{c}{AUROC} & \multicolumn{1}{c|}{AUPRC} & \multicolumn{1}{c}{Sensitivity} & \multicolumn{1}{c}{Specificity} & \multicolumn{1}{c}{Youden} & \multicolumn{1}{c}{AUROC} & \multicolumn{1}{c}{AUPRC}\\
          \hline
          \multirow{5}[2]{*}{POD1} & classifier single & 0.7553 & \textbf{0.7289} & 0.4842 & 0.7409 & 0.8068 & 0.7711 & \textbf{0.7737} & \textbf{0.5447} & 0.7468 & \underline{0.8293} \\
          & Informer & 0.8000 & 0.4550 & 0.2550 & 0.6840 & 0.7474 & 0.8550 & 0.5100 & 0.3650 & 0.7204 & 0.7816 \\
          & iTransformer & 0.9350 & 0.6850 & \textbf{0.6200} & 0.8589 & \underline{0.8578} & \underline{0.9350} & 0.5100 & 0.4450 & \underline{0.8644} & 0.8118 \\
          & Crossformer & 0.5400 & \underline{0.7000} & 0.2400 & 0.7292 & 0.7064 & 0.6850 & \underline{0.6550} & 0.3400 & \underline{0.8644} & 0.7538 \\
          & Origin Pathformer & \underline{0.9650} & 0.6250 & 0.5900 & \textbf{0.9282} & 0.8513 & \textbf{1.0000} & 0.5400 & \underline{0.5400} & \textbf{0.9220} & \textbf{0.8425} \\
          & Fusion Pathformer(ours) & \textbf{1.0000} & 0.6100 & \underline{0.6100} & \underline{0.9238} & \textbf{0.8597} & 0.9300 & 0.5500 & 0.4800 & 0.6784 & 0.8195 \\
          \hline
          \multirow{5}[2]{*}{POD2} & classifier single & 0.7395 & 0.6263 & 0.3658 & 0.7353 & 0.7670 & 0.7263 & 0.6447 & 0.3711 & 0.7485 & 0.7673 \\
          & Informer & \textbf{1.0000} & 0.7100 & \underline{0.7100} & 0.8607 & \underline{0.8876} & \textbf{1.0000} & 0.4600 & 0.4600 & 0.8319 & 0.8247 \\
          & iTransformer & 0.9350 & 0.6200 & 0.5550 & 0.9176 & 0.8393 & 0.9350 & 0.4800 & 0.4150 & 0.8599 & 0.8051 \\
          & Crossformer & 0.5750 & \textbf{0.7600} & 0.3350 & 0.8046 & 0.7465 & 0.5750 & \textbf{0.8050} & 0.3800 & 0.8079 & 0.7671 \\
          & Origin Pathformer & \underline{0.9450} & 0.7050 & 0.6500 & \underline{0.9384} & 0.8673 & \underline{0.9450} & 0.6500 & \underline{0.5950} & \underline{0.9274} & \underline{0.8511} \\
          & Fusion Pathformer(ours) & \textbf{1.0000} & \underline{0.7450} & \textbf{0.7450} & \textbf{0.9905} & \textbf{0.8984} & 0.9300 & \underline{0.7450} & \textbf{0.6750} & \textbf{0.9340} & \textbf{0.8749} \\
          \hline
          \multirow{5}[2]{*}{POD3} & classifier single & 0.6711 & 0.5711 & 0.2421 & 0.7061 & 0.7228 & 0.6395 & 0.7342 & 0.3737 & 0.7027 & 0.7631 \\
          & Informer & \textbf{1.0000} & 0.6400 & 0.6400 & \textbf{0.9859} & 0.8676 & \textbf{0.9950} & 0.6400 & 0.6350 & \textbf{0.9659} & 0.8659 \\
          & iTransformer & 0.9150 & \underline{0.8450} & \underline{0.7600} & \underline{0.9530} & \textbf{0.9063} & \underline{0.9750} & 0.6150 & 0.5900 & 0.9240 & 0.8522 \\
          & Crossformer & 0.5550 & \textbf{0.8500} & 0.4050 & 0.7901 & 0.7824 & 0.6100 & \textbf{0.9150} & 0.5250 & 0.8121 & 0.8413 \\
          & Origin Pathformer & 0.9400 & 0.8100 & 0.7500 & 0.9472 & 0.9009 & \underline{0.9750} & 0.7350 & \textbf{0.7100} & \underline{0.9451} & \textbf{0.8869} \\
          & Fusion Pathformer(ours) & \underline{0.9450} & 0.8200 & \textbf{0.7650} & 0.9438 & \underline{0.9062} & 0.9300 & \underline{0.7500} & \underline{0.6800} & 0.9304 & \underline{0.8769} \\
          \hline
      \end{tabular}
      }
      \caption{\textbf{Detailed experiment results on patient TYPE I.} The best results are highlighted in bold across different models, and the second-
best results are underlined.}
    \label{tab:patient1}
\end{table}

\begin{table}[!ht] 
      \centering
      \resizebox{\textwidth}{!}{  
      \begin{tabular}{c|c|ccccc|ccccc}
            \hline
            \multicolumn{1}{c|}{\multirow{2}{*}{Indicator}} & \multicolumn{1}{c|}{\multirow{2}{*}{Model}} & \multicolumn{5}{c}{Logistic Regression} & \multicolumn{5}{c}{SVM} \\
            \cline{3-12}
            & & \multicolumn{1}{c}{Sensitivity} & \multicolumn{1}{c}{Specificity} & \multicolumn{1}{c}{Youden} & \multicolumn{1}{c}{AUROC} & \multicolumn{1}{c|}{AUPRC} & \multicolumn{1}{c}{Sensitivity} & \multicolumn{1}{c}{Specificity} & \multicolumn{1}{c}{Youden} & \multicolumn{1}{c}{AUROC} & \multicolumn{1}{c}{AUPRC} \\
            \hline
            \multirow{5}[2]{*}{POD1} & classifier single & 0.7553 & \textbf{0.7289} & 0.4842 & 0.7409 & 0.8068 & 0.7711 & \textbf{0.7737} & \underline{0.5447} & 0.7468 & 0.8293 \\
            & Origin without TrendLoss & 0.9550 & \underline{0.6300} & 0.5850 & \underline{0.9276} & 0.8491 & \underline{0.9750} & \underline{0.6250} & \textbf{0.6000} & \underline{0.8869} & \textbf{0.8549} \\
            & Origin with TrendLoss & \underline{0.9650} & 0.6250 & \underline{0.5900} & \textbf{0.9282} & \underline{0.8513} & \textbf{1.0000} & 0.5400 & 0.5400 & \textbf{0.9220} & \underline{0.8425} \\
            & Fusion without TrendLoss & 0.9500 & 0.5650 & 0.5150 & 0.8865 & 0.8305 & 0.9250 & 0.5100 & 0.4350 & 0.6993 & 0.8081 \\
            & Fusion with TrendLoss & \textbf{1.0000} & 0.6100 & \textbf{0.6100} & 0.9238 & \textbf{0.8597} & 0.9300 & 0.5500 & 0.4800 & 0.6784 & 0.8195 \\
            \hline
            \multirow{5}[2]{*}{POD2} & classifier single & 0.7395 & 0.6263 & 0.3658 & 0.7353 & 0.7670 & 0.7263 & 0.6447 & 0.3711 & 0.7485 & 0.7673 \\
            & Origin without TrendLoss & 0.9550 & 0.6550 & 0.6100 & 0.9337 & 0.8561 & \underline{0.9350} & \underline{0.7150} & \underline{0.6500} & 0.9178 & \underline{0.8669} \\
            & Origin with TrendLoss & 0.9450 & \underline{0.7050} & \underline{0.6500} & 0.9384 & \underline{0.8673} & \textbf{0.9450} & 0.6500 & 0.5950 & 0.9274 & 0.8511 \\
            & Fusion without TrendLoss & \underline{0.9700} & 0.6650 & 0.6350 & \underline{0.9718} & 0.8641 & 0.9300 & \textbf{0.7450} & \textbf{0.6750} & \textbf{0.9342} & \textbf{0.8749} \\
            & Fusion with TrendLoss & \textbf{1.0000} & \textbf{0.7450} & \textbf{0.7450} & \textbf{0.9905} & \textbf{0.8984} & 0.9300 & \textbf{0.7450} & \textbf{0.6750} & \underline{0.9340} & \textbf{0.8749} \\
            \hline
            \multirow{5}[2]{*}{POD3} & classifier single & 0.6711 & 0.5711 & 0.2421 & 0.7061 & 0.7228 & 0.6395 & 0.7342 & 0.3737 & 0.7027 & 0.7631 \\
            & Origin without TrendLoss & 0.9350 & 0.7750 & 0.7100 & 0.9351 & 0.8868 & \textbf{0.9750} & \textbf{0.8050} & \textbf{0.7800} & \underline{0.9593} & \textbf{0.9104} \\
            & Origin with TrendLoss & 0.9400 & \underline{0.8100} & \underline{0.7500} & \underline{0.9472} & \underline{0.9009} & \textbf{0.9750} & 0.7350 & 0.7100 & 0.9451 & 0.8869 \\
            & Fusion without TrendLoss & \textbf{0.9700} & 0.7600 & 0.7300 & \textbf{0.9531} & 0.8933 & \underline{0.9600} & \underline{0.7800} & \underline{0.7400} & \textbf{0.9700} & \underline{0.8968} \\
            & Fusion with TrendLoss & \underline{0.9450} & \textbf{0.8200} & \textbf{0.7650} & 0.9438 & \textbf{0.9062} & 0.9300 & 0.7500 & 0.6900 & 0.9304 & 0.8766 \\
            \hline
      \end{tabular}
      }
      \caption{\textbf{Detailed ablation experiment results on patient TYPE I.} The best results are highlighted in bold across different models, and the second-
best results are underlined.
}
\label{tab:ablation}
\end{table}

\begin{table}[!ht] 
      \centering
      \resizebox{\textwidth}{!}{  
  \begin{tabular}{c|c|ccccc|ccccc}
      \hline
      \multicolumn{1}{c|}{\multirow{2}{*}{Indicator}} & \multicolumn{1}{c|}{\multirow{2}{*}{Model}} & \multicolumn{5}{c}{Logistic Regression} & \multicolumn{5}{c}{SVM} \\
      \cline{3-12}
      & & \multicolumn{1}{c}{Sensitivity} & \multicolumn{1}{c}{Specificity} & \multicolumn{1}{c}{Youden} & \multicolumn{1}{c}{AUROC} & \multicolumn{1}{c|}{AUPRC} & \multicolumn{1}{c}{Sensitivity} & \multicolumn{1}{c}{Specificity} & \multicolumn{1}{c}{Youden} & \multicolumn{1}{c}{AUROC} & \multicolumn{1}{c}{AUPRC}\\
      \hline
      \multirow{8}[2]{*}{POD1} & classifier single & \textbf{0.4216} & 0.6660 & \underline{0.0877} & 0.4653 & \underline{0.6344} & 0.3845 & 0.6839 & 0.0685 & 0.4594 & 0.6206 \\
      & Transformer & \underline{0.4102} & \underline{0.7725} & \textbf{0.1826} & \textbf{0.6520} & \textbf{0.6741} & \textbf{0.5459} & 0.6123 & \textbf{0.1582} & \textbf{0.6475} & \textbf{0.6788} \\
      & Autoformer & 0.2266 & \textbf{0.7979} & 0.0244 & \underline{0.6071} & 0.5709 & 0.3545 & \textbf{0.7852} & \underline{0.1396} & \underline{0.6310} & \underline{0.6499} \\
      & Informer & 0.3799 & 0.5869 & -0.0332 & 0.4719 & 0.5845 & \underline{0.4209} & 0.6787 & 0.0996 & 0.5959 & 0.6388 \\
      & iTransformer & 0.2979 & 0.7236 & 0.0215 & 0.4869 & 0.5838 & 0.2969 & \underline{0.7393} & 0.0361 & 0.4925 & 0.5904 \\
      & Crossformer & 0.1787 & 0.6367 & -0.1846 & 0.4091 & 0.4595 & 0.2119 & 0.6943 & -0.0938 & 0.4003 & 0.5077 \\
      & Origin Pathformer & 0.3213 & 0.6982 & 0.0195 & 0.4699 & 0.5882 & 0.3555 & 0.6328 & -0.0117 & 0.4850 & 0.5848 \\
      & Fusion Pathformer(ours) & 0.2363 & 0.7246 & -0.0391 & 0.5172 & 0.5400 & 0.3662 & 0.7012 & 0.0674 & 0.5364 & 0.6169 \\
      \hline
      \multirow{8}[2]{*}{POD2} & classifier single & 0.2418 & 0.6276 & -0.1305 & 0.3806 & 0.5073 & 0.2790 & 0.5393 & -0.1817 &0.3788 & 0.5083 \\
      & Transformer & 0.2432 & \underline{0.8662} & \underline{0.1094} & \textbf{0.7086} & \underline{0.6333} & \underline{0.3789} & \textbf{0.9395} & \underline{0.3184} & \underline{0.6881} & \textbf{0.7758} \\
      & Autoformer & 0.1406 & \textbf{0.8867} & 0.0273 & \underline{0.6311} & 0.5621 & 0.1699 & 0.8486 & 0.0186 & 0.588 & 0.5569 \\
      & Informer & \textbf{0.4014} & 0.5762 & -0.0225 & 0.5061 & 0.5935 & \textbf{0.4775} & \underline{0.8867} & \textbf{0.3643} & \textbf{0.7707} & \underline{0.7735} \\
      & iTransformer & 0.1787 & 0.7520 & -0.0693 & 0.5011 & 0.5041 & 0.1855 & 0.7148 & -0.0996 & 0.4942 & 0.4932 \\
      & Crossformer & \underline{0.3809} & 0.7998 & \textbf{0.1807} & 0.5919 & \textbf{0.6729} & 0.2422 & 0.7354 & -0.0225 & 0.5224 & 0.5495 \\	
      & Origin Pathformer & 0.1982 & 0.7637 & -0.0381 & 0.5549 & 0.5277 & 0.2109 & 0.6729 & -0.1162 & 0.5145 & 0.4987 \\
      & Fusion Pathformer(ours) & 0.1494 & 0.7626 & -0.0879 & 0.5723 & 0.4805 & 0.1689 & 0.7402 & -0.0908 & 0.5652 & 0.4893 \\
      \hline
      \multirow{8}[2]{*}{POD3} & classifier single & 0.1185 & 0.6807 & -0.2008 & 0.4161 & 0.4150 & 0.1185 & 0.6922 & -0.1893 & 0.3873 & 0.4187 \\
      & Transformer & 0.3459 & \textbf{0.9738} & 0.3198 & \textbf{0.8306} & \underline{0.8013} & 0.4506 & \textbf{0.9448} & \textbf{0.3953} & \textbf{0.8622} & \textbf{0.8080} \\
      & Autoformer & \underline{0.4884} & \underline{0.9244} & \textbf{0.4128} & \underline{0.7026} & \textbf{0.8051} & \textbf{0.5058} & \underline{0.8750} & \underline{0.3808} & 0.6933 & \underline{0.7774} \\
      & Informer & \textbf{0.6715} & 0.6744 & \underline{0.3459} & 0.7021 & 0.7546 & 0.5378 & 0.8343 & 0.3721 & \underline{0.7880} & 0.7667 \\
      & iTransformer & 0.1250 & 0.8721 & -0.0029 & 0.4888 & 0.5284 & 0.1744 & 0.8256 & 0.0000 & 0.5068 & 0.5436 \\
      & Crossformer & 0.0552 & 0.8256 & -0.1192 & 0.4951 & 0.3841 & 0.0262 & 0.8023 & -0.1715 & 0.4252 & 0.3150 \\
      & Origin Pathformer & 0.1395 & 0.8547 & -0.0058 & 0.5441 & 0.5298 & 0.1802 & 0.8110 & -0.0087 & 0.5454 & 0.5392 \\
      & Fusion Pathformer(ours) & 0.0523 & 0.8779 & -0.0698 & 0.6087 & 0.4131 & 0.2442 & 0.7762 & 0.0203 & 0.5387 & 0.5719 \\
      \hline
  \end{tabular}
}
\caption{\textbf{Detailed experiment results on patient TYPE II.} The best results are highlighted in bold across different models, and the second-
best results are underlined.}
\label{tab:patient2}
\end{table}

\end{document}